\documentclass[journal]{IEEEtran}
\usepackage{cite}
\ifCLASSINFOpdf
  \usepackage[pdftex]{graphicx}
\else
\fi
\usepackage{amsmath}
\usepackage{algorithmic}
\usepackage{array}
\usepackage{subfigure}
\usepackage{fixltx2e}
\usepackage{dblfloatfix}
\usepackage{url}
\usepackage{amsmath} 

\begin{document}
\title{Enhanced Sign Language Translation between American Sign Language (ASL) and Indian Sign Language (ISL) Using LLMs}

\author{Malay~Kumar,~S.~Sarvajit~Visagan,~Tanish~Sarang~Mahajan,~and~Anisha~Natarajan
%\thanks{Dr.~Anisha~Natarajan is with the School of Electronics Engineering, Vellore Institute of Technology, Chennai, TN, 600127, India (e-mail: anisha.natarajan@vit.ac.in).}%
% <-this % stops a space
%\thanks{Manuscript received April 19, 2024; revised August 26, 2015.}
}

\newcommand{\customcaption}[1]{\par\centering\small #1\par}

\maketitle

\begin{abstract}
We have come up with a research that hopes to provide a bridge between the users of American Sign Language and the users of spoken language and Indian Sign Language (ISL). The research enabled us to create a novel framework that we have developed for Learner Systems. Leveraging art of Large models to create key features including: - Real-time translation between these two sign languages in an efficient manner. Making LLM's capability available for seamless translations to ISL. Here is the full study showing its implementation in this paper. The core of the system is a sophisticated pipeline that begins with reclassification and recognition of ASL gestures based on a strong Random Forest Classifier. By recognizing the ASL, it is translated into text which can be more easily processed. Highly evolved natural language
NLP (Natural Language Processing) techniques come in handy as they play a role in our LLM integration where you then use LLMs to be able to convert the ASL text to ISL which provides you with the intent of sentence or phrase.
The final step is to synthesize the translated text back into ISL gestures, creating an end-to-end translation experience using RIFE-Net. This framework is tasked with key challenges such as automatically dealing with gesture variability and overcoming the linguistic differences between ASL and ISL. By automating the translation process, we hope to vastly improve accessibility for sign language users. No longer will the communication gap between ASL and ISL create barriers; this totally cool innovation aims to bring our communities closer together. And we believe, with full confidence in our framework, that we're able to apply the same principles across a wide variety of sign language dialects.

\end{abstract}

\begin{IEEEkeywords}
Sign Language Recognition,
ASL to ISL Translation,
Large Language Models (LLMs),
Natural Language Processing (NLP),
Random Forest Classifier,
Gesture Reclassification,
Text-to-Gesture Synthesis,
RIFE-Net,
Real-Time Translation,
Sign Language Variability,
Linguistic Adaptation,
Assistive Technology,
Cross-Linguistic Framework,
Sign Language Dialects,
Accessibility and Inclusivity
\end{IEEEkeywords}

% \fi
\IEEEpeerreviewmaketitle

\section{Introduction}
\IEEEPARstart{T}{he} communication gap between the users of American Sign Language (ASL) and Indian Sign Language (ISL)
is a significant challenge to intercultural interaction and
accessibility in the deaf community. Sign languages are
indispensable weapons of expression for deaf individuals,
but the lack of interoperability between ASL and ISL limits
smooth communication across various linguistic and cultural frontiers. This challenge can be addressed through
innovative strategies that leverage advances in machine
translation technology and contemporary deep learning
frameworks.
The central goal of this research work is to build a
holistic framework for machine translation that can efficiently and conveniently translate ASL gestures into ISL gestures, hence facilitating communication between ASL users and ISL users. Toward this objective, we hereby propose
an approach that combines image recognition techniques
with advanced language processing algorithms governed by
Large Language Models. Instead of using traditional CNN-
based recognition techniques, LLM-driven techniques are
used to decode ASL gestures and, in turn, translate directly
into meaningful textual representations.
From this step of converting ASL gestures to text, we establish an intermediate that helps use advanced LLM-based
techniques for machine translation. Now, we can translate
the recognized English text into ISL gestures by preserving
linguistic aspects and cultural context. The application of
LLMs for translation will help make the process more
accurate, context-sensitive, and adaptable and, thus, would
be used to bridge between ASL and ISL in a nearly seamless
manner.

\subsection{Leveraging Deep Learning for Gesture Recognition and Translation}
Sign language communication involves complex expressions that carry a certain range of linguistic and cultural nuances in communication. Conventional sign language understanding and translation techniques rely mostly on fixed data sets and prebuilt models. Such approaches find it difficult to adapt to the dynamic nature of sign languages. In the paper, the constraints were overcome using Random Forest Classifiers for effective gesture recognition followed by Large Language Models to assist in context-aware translation. The proposed framework, therefore, would represent a breakthrough with the integration of real-time processing capabilities and cultural contextualization.

The significance of this work is that it introduces an intermediate text-based representation that forms the connecting link between recognition and synthesis of gestures. This linguistic intermediate allows for not only a more literal translation but also allows the tailoring of the translation method in such a way as to keep its intent and cultural nuances of the original expression. This model would become especially important to the reduction of the linguistic differences between ASL and ISL, such as differing grammatical structures, word orders, and contextual expressions. Further, generating ISL gestures from translated text with the assistance of RIFE-Net produces a smooth and natural gesture presentation. Such a robust ability of RIFE-Net to effectively handle high variability within gesture sequences leads it to accurately reproduce ISL gestures even while making forward-looking predictions of potentially complex or subtle translations. With advanced recognition, translation, and synthesis modules integrated, the framework places it at the forefront of a cutting-edge tool in the domain of sign language translation.

%\IEEEpubidadjcol

\subsection{Towards a Multi-dialectal Future}
Though designed specifically for ASL and ISL, underpinning principles and methodologies are adaptable to support more than those targeted sign language dialects. By adapting the system to use a variety of datasets, it follows that it is viable to address the diversity in sign languages globally, thereby paving the course for universal sign language interoperability. Moreover, this adaptability brings out the scale in using LLMs for translation purposes as it opens up broader applications in assistive technologies.

This research promises to work out an integrated approach combining gesture recognition, natural language processing, and synthesis of gestures to bridge the overall communication gap so present in sign language communities. This is an advance both in the technical and importance toward furthering the aspects of inclusivity and exchange across the world's deaf community.

\section{Literature Survey}
Recognition and translation of sign languages have attracted a lot of focus in the recent past because of their pivotal role in improving communication between deaf and mute people. Presenting an overview of the literature in this field, this literature review identifies methodologies, techniques, and challenges in sign language recognition and translation systems.
 
Rao et al. (2023) introduced a system of how people with many spoken languages utter sign language using the Natural Language Toolkit (NLTK) and argued that a critical factor in sign language translation is linguistic processing \cite{Rao2023}. A popular review carried out by Al-Qureshi et al. (2023) focused on the DP of recognizing sign languages through deep learning approaches, including challenges and recent milestones \cite{AlQureshi2021}. Salian et al. (2017) also presented a system for sign language recognition to set the background for subsequent work in this domain \cite{Salian2017}.
 
In this Article, Mubashira and James 2020 proposed a Transformer Network for video-to-text translation proving the transformers effective in sign language \cite{Mubashira2020}. Halder and Tayade (2021) in a recent study that designed a real-time vernacular sign language recognition system using media pipe and machine learning propose the possibility of real-time applications in this field \cite{Halder2021}. Also, in Korean sign language recognition, Shin et al. (2023) proposed a Transformer-Based Deep Neural Network and demonstrated how deep learning can be applied for sign language recognition irrespective of the type of sign language \cite{Shin2023}.
 
Saleem et al. (2023) provided an outstanding, machine learning-based, full-duplex sign language communication system capable of translating multiple sign languages; the problem of sign language communication technology for the deaf community has received continued attention \cite{Saleem2023}. Wang et al. (2022) proposed an improved 3D-Res-Net sign language recognition algorithm with clearer hand features and improved the state of the art of gesture recognition \cite{Wang2022}.
 
In their paper, Sahoo et al. (2014) gave a detailed account of the recent approaches to recognizing sign languages, and major issues \cite{Sahoo2014}. Pathan et al. Pathan, Muhammad, and Zhang (2023) developed a fusion approach for sign language recognition from both image and hand landmarks using a multi-headed convolutional neural network demonstrating how multiple-modal can enhance gesture recognition systems \cite{Pathan2023}.
 
Shenoy et al. (2018) have proposed real-time ISL recognition using grid-based feature extraction and machine learning to address the requirements of the Indian deaf and Hard of hearing community \cite{Shenoy2018}. Sharma and Singh (2022) investigated NLP in the general mapping of speech to ISL with insights towards a hybrid way of sign language translation \cite{Sharma2022}.
 
In addition, while Badhe and Kulkarni (2015) developed an Indian sign language translator based on gesture recognition algorithms, they pointed out that feature extraction and motion detection remain decisive factors in designing sign language recognition systems \cite{Badhe2015}.
 
Furthermore, Jia Gong (2024), the author was motivated by the remarkable translation performance of LLMs and proposed a method to incorporate off-the-shelf LLMs to address complex Sign Language Translation Tasks \cite{Gong2024}. ZhiGang Chen (2024) thought exploring the gloss-free methods is much more crucial because it will greatly decrease the annotation time and promote the dependence of more accurate and universal sign language translation systems \cite{Chen2024}. Sen Fang then unveiled SIGNLLM which is the first large-scale multilingual SLP model created using the prompt2 sign dataset which generates the skeletal poses of sign language from text or prompt for eight languages \cite{Fang2024}.
\section{Proposed Methodology}
This research presents a novel methodology to bridge the communication gap between American Sign Language (ASL) and Indian Sign Language (ISL) through an automated translation system enhanced using Large language models (LLMs). The proposed approach comprises three interconnected phases: The sign language recognition phase that utilizes a hybrid ensemble model, the Recognized Text Correction phase through language model enhancement, and the Video synthesis phase with motion smoothing. Each phase has been meticulously designed to address specific challenges in cross-sign-language translation while maintaining semantic accuracy and natural gesture flow.
\subsection{Sign Language Recognition}
The initial phase employs a hybrid ensemble approach combining the complementary strengths of a Random Forest Classifier (RFC) and a Convolutional Neural Network (CNN). This RFC+CNN model architecture allows robust feature extraction and classification of the sign language and provides better prediction accuracy.
\subsubsection{Random Forest Classifier Model}
The Random Forest Classifier component focuses on the skeletal approach which sets a baseline for sign language recognition, it uses the hand-tracking capabilities of the mediapipe framework for feature extraction.
\\
Dataset Preparation: For the above model we prepare an image dataset with a sample size of 2800 images (28 distinct classes * 100 images)
\begin{itemize}
    \item 26 alphabetic characters (100 images per character).
    \item Additional classes for 'DELETE' and 'SPACE' operations.
    \item A total of 28 distinct classes.
\end{itemize}
\begin{table}[ht]
    \centering
    \caption{Dataset Distribution and Acquisition Parameters}
    \label{tab:dataset_params}
    \begin{tabular}{p{1.5cm} p{3cm} p{3cm}}
        \hline
        \textbf{Parameter} & \textbf{Specification} & \textbf{Justification} \\ \hline\hline
        Image Resolution & 32×32 pixels & Optimal balance between detail preservation and computational efficiency \\ \hline
        Lighting Conditions & Ambient, Direct, Shadow & Ensures robustness across various lighting scenarios \\ \hline
        Background Variation & Background clutter, objects & Improves model generalization \\ \hline
        Camera Angles & 0°, ±15°, ±30° & Accounts for natural variation in hand positioning \\ \hline
        Crop Focus & Hand region & Eliminates irrelevant background information \\ \hline
    \end{tabular}
\end{table}
\begin{figure}[h]
    \centering
    \includegraphics[width=1.0\linewidth]{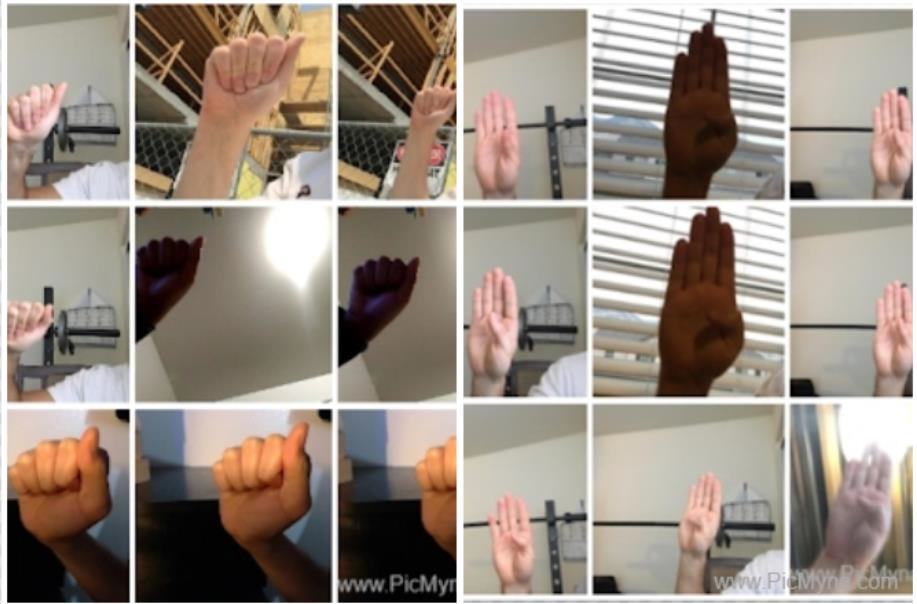}
    \caption{American Sign Language dataset sample}
    \label{fig:rfc-asl-dataset}
\end{figure}
Feature extraction: With the mediapipe framework we extract 42 prominent landmark points (L) placed on the hand which serve as the features for training the model.
\[
L = \{(x_i, y_i, z_i) \mid i \in [1, 42]\}
\]
% \begin{figure}[h]
%     \centering
%     \includegraphics[width=1.0\linewidth]{Images/feature_importance_rfc.png}
%     \caption{Feature Importance - Hand Landmarks (42)}
%     \label{fig:fea_imp_rfc}
% \end{figure}
The coordinates obtained are then processed for normalization using StandardScaler.
\[
x_{\text{normalized}} = \frac{x - \mu_x}{\sigma_x}
\]
where $\mu_x$ and $\sigma_x$ represent the mean and standard deviation of the feature distribution respectively.
\\\\
Model Architecture and Training: We use the RandomForestClassifer (RFC) model that employs an optimized through the GridSearch Algorithm.
\begin{table}[ht]
\centering
    \caption{RFC Hyperparameter Configuration}
\label{tab:rfc_hyperparams}
\begin{tabular}{p{2cm} p{2cm} p{1cm} p{2.5cm}}
\hline
\textbf{Parameter} & \textbf{Search Space} & \textbf{Best Value} & \textbf{Impact} \\ \hline \hline
\textbf{n\_estimators}       & {[}100, 200, 300{]} & 200  & \begin{tabular}[c]{@{}l@{}}Balances complexity\\ and accuracy\end{tabular} \\
\textbf{max\_depth}         & {[}None, 10, 20, 30{]} & 20                  & Controls overfitting \\
\textbf{min\_samples\_split} & {[}2, 5, 10{]}      & 5    & \begin{tabular}[c]{@{}l@{}}Ensures robust\\  node splitting\end{tabular}   \\
\textbf{min\_samples\_leaf} & {[}1, 2, 4{]}          & 2                   & Prediction stability \\
\textbf{bootstrap}           & {[}True, False{]}   & True & \begin{tabular}[c]{@{}l@{}}Enables varied tree\\ construction\end{tabular} \\ \hline
\end{tabular}
\end{table}
\\
The model's decision function for class prediction can be expressed as:
\[\hat{y} = \text{mode }{\hat{y}t(x)}{t=1}^T\]
where $\hat{y}_t(x)$ represents the prediction of the t-th tree, and T is the total number of trees.
\\
\subsubsection{Convolutional Neural Network Model}
For the convolutional neural network (CNN) component we use silhouette images providing complementary visual feature analysis to the RFC's landmark-based approach.
\\
Silhouette Images: It is a type of image where a dark image of a subject against a lighter background, usually showing the subject's profile.
% \begin{figure}[ht]
%     \centering
%     \includegraphics[width=0.5\linewidth]{Images/sample_silhoutte_image.png}
%     \caption{Sample Silhouette Image}
%     \label{fig:sample-silhoutte-image}
% \end{figure}
\\
Using silhouette images enables a focus on the intrinsic structure of the hand pose, effectively eliminating background noise, lighting variations, and other environmental influences. This approach isolates the key features of the hand's position and configuration, providing a clearer and more accurate representation for training the model.
\\
Dataset Preparation: For the CNN model we generate silhouette images of the American sign language (ASL) with a sample size of 16200 images (27 distinct classes * 600 images).
\begin{itemize}
    \item 26 alphabetic characters (600 images per character)
    \item Additional class for "BLANK" image which serves as ground truth.
    \item Augmentation of images:
    \begin{itemize}
        \item Horizontal flipping: $I_{flip}(x,y) = I(w-x,y)$
        \item Brightness variation: $I_{bright}(x,y) = \alpha I(x,y)$, where $\alpha \in [0.8, 1.2]$
        \item Gaussian noise addition: $I_{noise}(x,y) = I(x,y) + \mathcal{N}(0,\sigma^2)$ \\
    \end{itemize}
\end{itemize}
\begin{figure}[ht]
    \centering
    \includegraphics[width=1\linewidth]{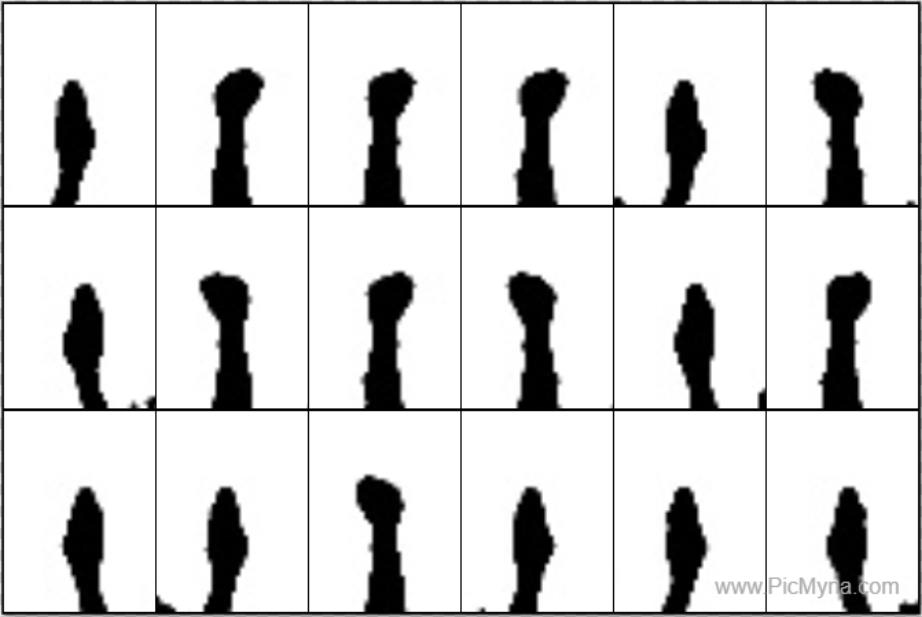}
    \caption{Silhoutte Image American Sign Language dataset}
    \label{fig:cnn-sil-dataset}
\end{figure}
Silhouette image generation:
In the experimental phase, we evaluated three distinct approaches for generating silhouette images: Histogram-based thresholding, Skin colour segmentation, and Otsu-thresholding.
\begin{figure}[h]
    \centering
    \subfigure[Hist-Thresholding]{%
        \includegraphics[width=0.3\linewidth]{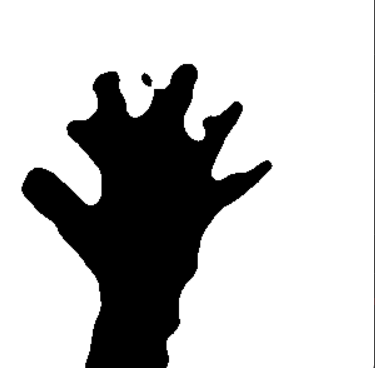}
        }
        \hfill
    \subfigure[Skin-Color Segmentation]{%
        \includegraphics[width=0.3\linewidth]{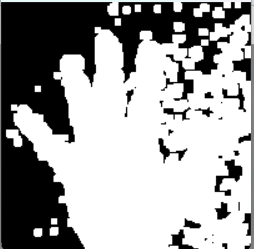}
        }
        \hfill
    \subfigure[Otsu-Thresholding]{%
        \includegraphics[width=0.3\linewidth]{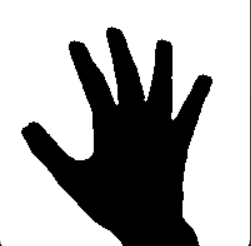}
        }
\end{figure}

Otsu-thresholding proved most effective
\[threshold = \text{argmax}_t{\sigma_B^2(t)}\]
where $\sigma_B^2(t)$ is the between-class variance:
\[\sigma_B^2(t) = \omega_0(t)\omega_1(t)[\mu_0(t) - \mu_1(t)]^2\]
\\
CNN Architecture:
\begin{table}[ht]
\centering
    \caption{CNN Layer Configuration and Specifications}
\label{tab:cnn_architecture}
\begin{tabular}{llll}
\hline
\multicolumn{1}{l}{\textbf{Layer}} & \multicolumn{1}{l}{\textbf{Configuration}} & \multicolumn{1}{l}{\textbf{Output Shape}} & \multicolumn{1}{l}{\textbf{Parameters}} \\ \hline \hline
Conv2D    & 16 filters, (2,2) kernel & (31,31,16)     & 80     \\
MaxPool2D & (2,2) pool, (2,2) stride & (15,15,16)     & 0      \\
Conv2D    & 32 filters, (3,3) kernel & (13,13,32)     & 4,640  \\
MaxPool2D & (3,3) pool, (3,3) stride & (4,4,32)       & 0      \\
Conv2D    & 64 filters, (5,5) kernel & (4,4,64)       & 51,264 \\
MaxPool2D & (5,5) pool, (5,5) stride & (1,1,64)       & 0      \\
Dense     & 128 units                & (128)          & 8,320  \\
Dropout   & 0.2 rate                 & (128)          & 0      \\
Dense     & num\_classes             & (num\_classes) & varies \\ \hline
\end{tabular}
\end{table}
Training specifications:
\begin{itemize}
    \item Loss function: Categorical Cross-entropy
    \[L = -\sum_{i=1}^{C} y_i \log(\hat{y}_i)\]
    \item Optimizer: Adam with learning rate 1e-3
    \item Batch size: 32
    \item Epochs: 100 with early stopping
\end{itemize}

Ensemble Integration:
The ensemble combines predictions through a weighted voting mechanism:
\[P(y|x) = w_{RFC}P_{RFC}(y|x) + w_{CNN}P_{CNN}(y|x) \]

where $w_{RFC}$ and $w_{CNN}$ are optimized weights determined through validation performance.

\subsection{Recognized Text Correction}
In this phase, we implement text correction on the recognized text obtained from the sign language recognition phase. We use a fine-tuned large language model (LLM) to auto-correct the recognized text.
\\
For fine-tuning the pre-trained large language model (LLM) we prepare a dataset containing 500 examples. The dataset is structured as a collection of text pairs, with each entry containing:
\begin{itemize}
    \item Input Text: A set of characters or words that may contain typos, scrambled letters, incomplete words, or some non-standard character sequences.
    \item Output Text: A corrected and meaningful interpretation of the input text. This output contains three variations of grammatically correct phrases, which could be a sentence, a noun phrase, or a meaningful fragment.
\end{itemize}

\begin{table}[ht]
\centering
    \caption{Fine-Tune Data Characteristics}
\label{tab:fine_tune_data}
\begin{tabular}{llll}
\hline
\textbf{Error Type}                                               & \textbf{Percentage} & \textbf{Example Input} & \textbf{Corrected Output} \\ \hline
\begin{tabular}[c]{@{}l@{}}Character \\ Substitution\end{tabular} & 35\%                & "hllo"                 & "hello"                   \\
\begin{tabular}[c]{@{}l@{}}Missing \\ Characters\end{tabular} & 25\% & "wrld"      & "world"     \\
\begin{tabular}[c]{@{}l@{}}Extra \\ Characters\end{tabular}   & 20\% & "helploo"    & "hello"     \\
Word Order                                                    & 20\% & "you thank" & "thank you" \\ \hline
\end{tabular}
\end{table}
Given that our use case is limited to text correction within a predefined context, we prioritized the utilization of lightweight models to ensure efficiency. Specifically, we evaluated Gemini-1.5 Flash, GPT-3.5, and LLaMA 2 (7B) after fine-tuning and conducting a comprehensive performance analysis. Following this evaluation, Gemini-1.5 Flash was selected as the optimal model due to its minimal inference time and low memory requirements, making it well-suited for our constrained computational environment.

\begin{table}[ht]
\centering
    \caption{LLM Comparison and Selection Criteria}
\label{tab:llm_comparison}
\begin{tabular}{llll}
\hline
\textbf{Criteria} & \textbf{\begin{tabular}[c]{@{}l@{}}Gemini-1.5 Flash\\ (Selected)\end{tabular}} & \textbf{GPT-3.5} & \textbf{LLaMA2 7B} \\ \hline \hline
Accuracy       & 94.2\%      & 96.8\%        & 95.5\%       \\
\textbf{Inference Time} & 15ms        & 45ms          & 25ms         \\
\textbf{Memory Usage}  & 2.0GB       & 4.0GB         & 3.0GB        \\
Fine-tuning    & Limited     & Not Available & Full Control \\
Custom Opt.    & Limited     & No            & Yes          \\
License        & Proprietary & Proprietary   & Open Source  \\ \hline
\end{tabular}
\end{table}

\subsection{Video Synthesis}
The final phase of the proposed methodology involves mapping the corrected text to Indian Sign Language (ISL) gestures and creating a fluid video sequence.
\\
The Indian Sign Language gesture dataset comprises visual manifestations of the 26 letters of the ISL alphabet in sequential order from 0 to 25. Images were captured against a black background, which helps create a high level of contrast and enhances the visibility of hand signs. To ensure uniformity, all images from the ISL dataset will be resized to 128x128 pixels before being used in the output generation process.
\\
To generate video outputs from the initial mapping of characters to their corresponding signs, we initially produced video frames at 1 FPS. To enhance the temporal smoothness and improve the viewing experience, we first duplicated the frames to achieve a 24 FPS output. For further refinement, we employed RIFE-Net for advanced frame interpolation, enabling the generation of high-quality video outputs at 60 FPS.
\begin{itemize}
    \item Initial Frame Duplication
\[F_{out} = {F_i | F_i = F_{\lfloor i/24 \rfloor}, i \in [0,24n-1]}\]
    \item RIFE Network Implementation 
    \begin{itemize}
        \item Flow Estimation:
        \[\mathbf{F}{t\rightarrow 0}, \mathbf{F}{t\rightarrow 1} = \text{FlowNet}(I_0, I_1, t)\]
        \item Context Extraction:
        \[C_0, C_1 = \text{ContextNet}(I_0, I_1)\]
        \item Frame Synthesis:
        \[I_t = \text{FusionNet}(I_0, I_1, \mathbf{F}{t\rightarrow 0}, \mathbf{F}{t\rightarrow 1}, C_0, C_1)\]
    \end{itemize}
\end{itemize}

\section{Proposed Workflow} 

\subsubsection{Sign Language Recognition Module}
ASL gestures are converted into text through live camera feeds. The workflow involves:
\begin{itemize}
    \item Preprocessing: Isolating hand movements to extract meaningful features.
    \item Model Analysis: An ensemble model, comprising a Convolutional Neural Network (CNN) for spatial pattern extraction and a Random Forest Classifier (RFC) for gesture classification, generates text outputs. Due to recognition challenges, the initial text may contain errors (e.g., "HELOLO WRLD").
\end{itemize}
\subsubsection{Text Correction Module}
The recognized text is refined for syntactical and contextual accuracy using the Gemini-1.5 flash, a highly optimized Large Language Model (LLM).
\begin{itemize}
    \item Functionality: Gemini-1.5 leverages a multilingual dataset to rectify inaccuracies, ensuring precise and fluent output (e.g., correcting "HELOLO WRLD" to "HELLO WORLD").
\end{itemize}
\subsubsection{Video Synthesis Module}
Corrected text is mapped to ISL gesture frames using a predefined algorithm to generate ISL videos.
\begin{itemize}
    \item Video Enhancement: RIFE-Net (Real-Time Intermediate Flow Estimation) smooths and interpolates frames, producing natural 60 FPS videos from initial 1 FPS sequences. The resulting ISL video offers enhanced clarity and usability, facilitating effective communication for ISL users.
\end{itemize}

\begin{figure}[h]
    \centering
    \includegraphics[width=0.85\linewidth]{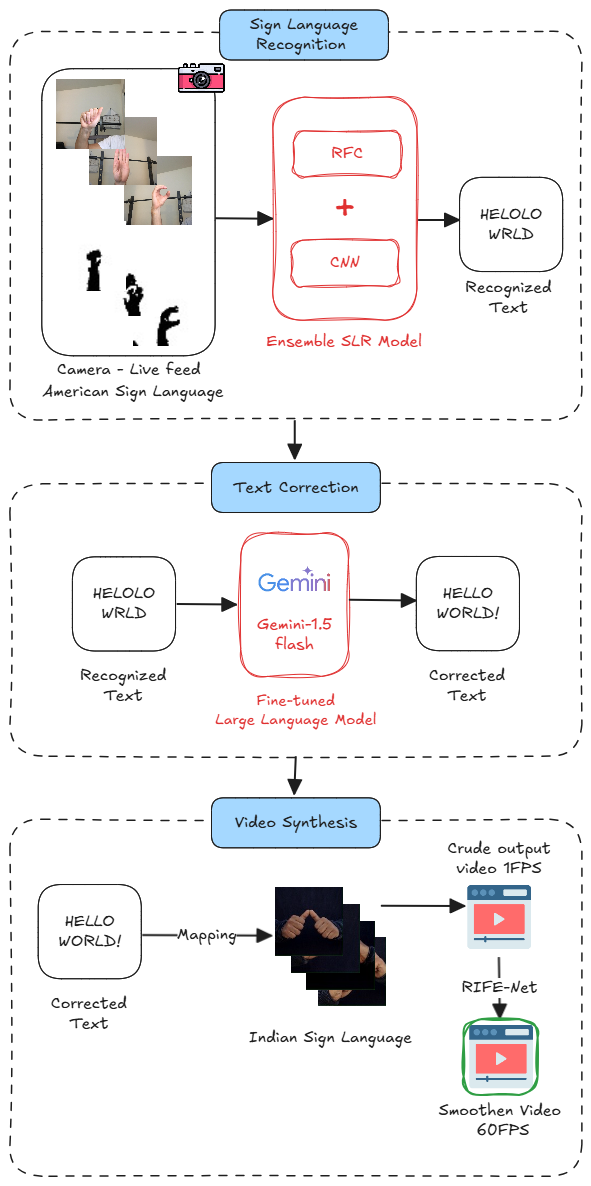}
    \caption{System Architecture}
    \label{fig:sys_arch}
\end{figure}

\section{Result and Output}
This section provides the experimental results of various phases of development, which are performed and investigated to build a complete framework. The proposed framework functionalities are tested in different stages of the development cycle. In addition to that, we have shown the user interface screens of the final application. For the first phase sign language recognition phase we implemented the hybrid approach RFC+CNN.
Below are the results obtained on training the RFC Model and the CNN Model.

\subsection{Random Forest Classifier Model}
The following is the confusion matrix obtained on testing the trained model on test sample set of 20 records for each class.
\begin{figure}[h]
    \centering
    \includegraphics[width=1\linewidth]{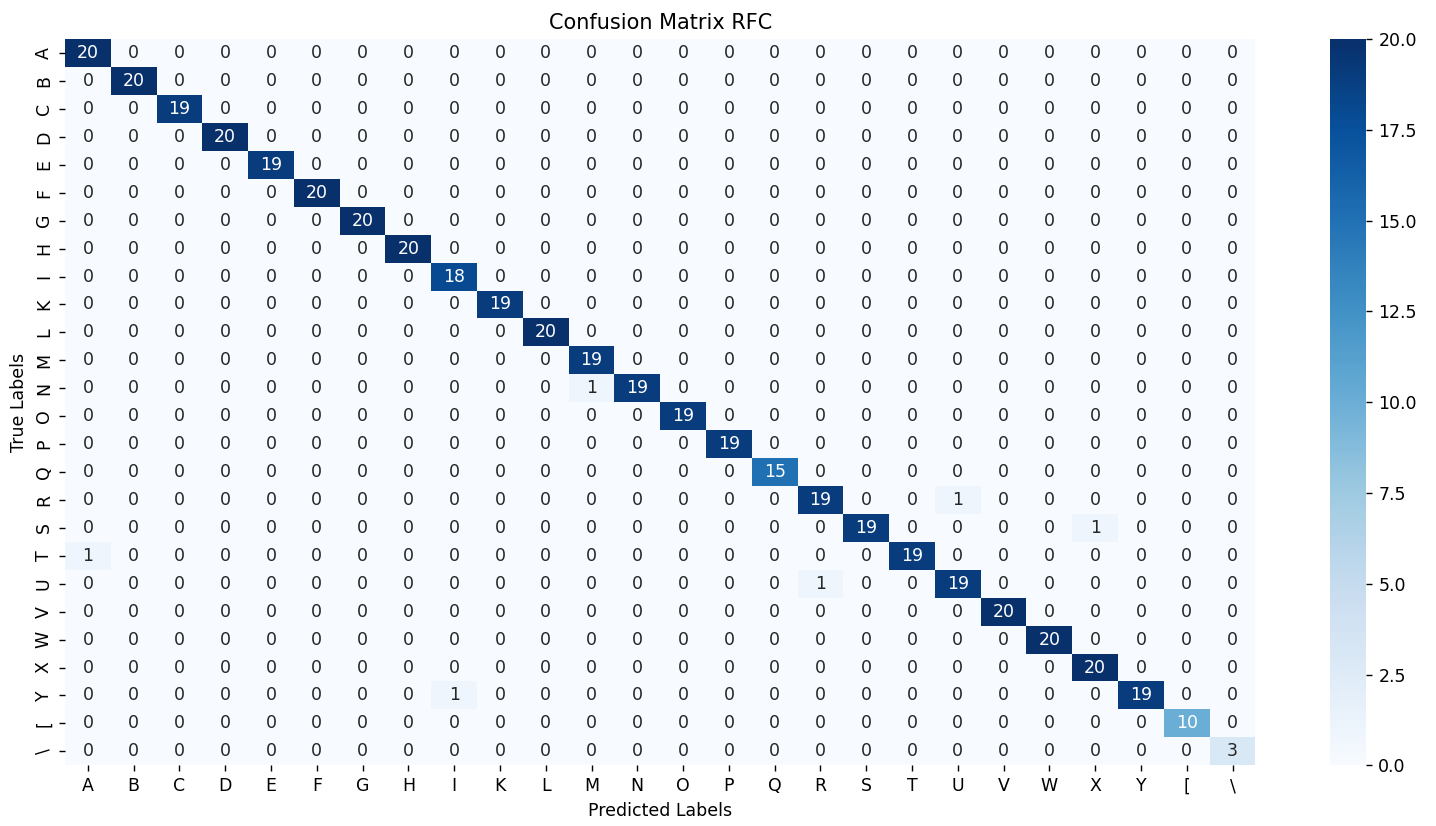}
    \caption{Confusion Matrix Random Forest Classifier Model}
    \label{fig:conf_matrix_rfc}
\end{figure}
\subsection{Convolutional Neural Networks Model}
On training the CNN model we obtain the following training evaluation graph for 15 epochs. Overall 82.4\% accuracy was obtained.
\begin{figure}[h]
    \centering
    \includegraphics[width=1\linewidth]{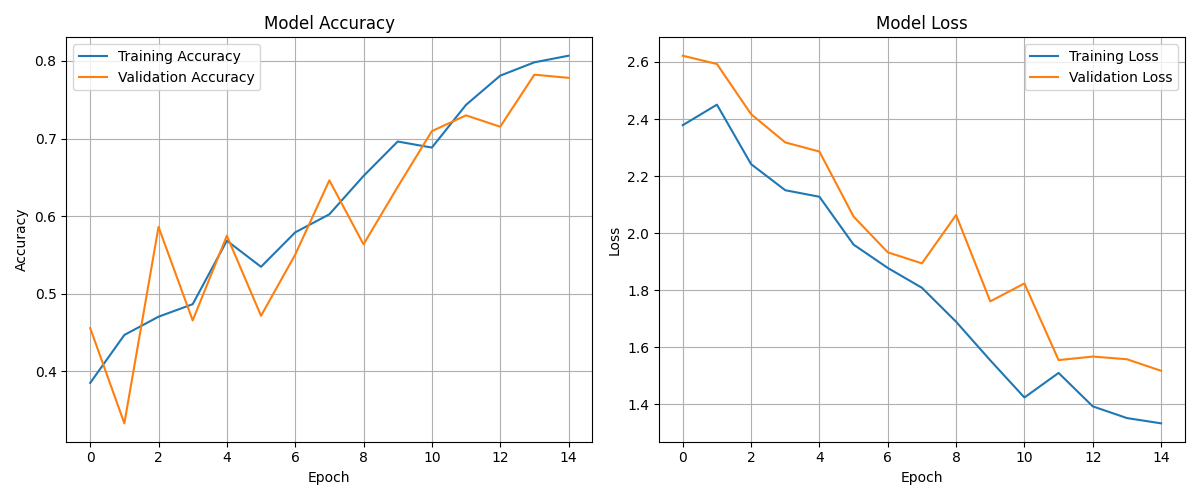}
    \caption{Accuracy and Loss Evaluation of the CNN Model}
    \label{fig:acc_loss_graphs}
\end{figure}
The following is the confusion matrix obtained on testing the trained model on test sample set of 100 records for each class.
\begin{figure}[h]
    \centering
    \includegraphics[width=1\linewidth]{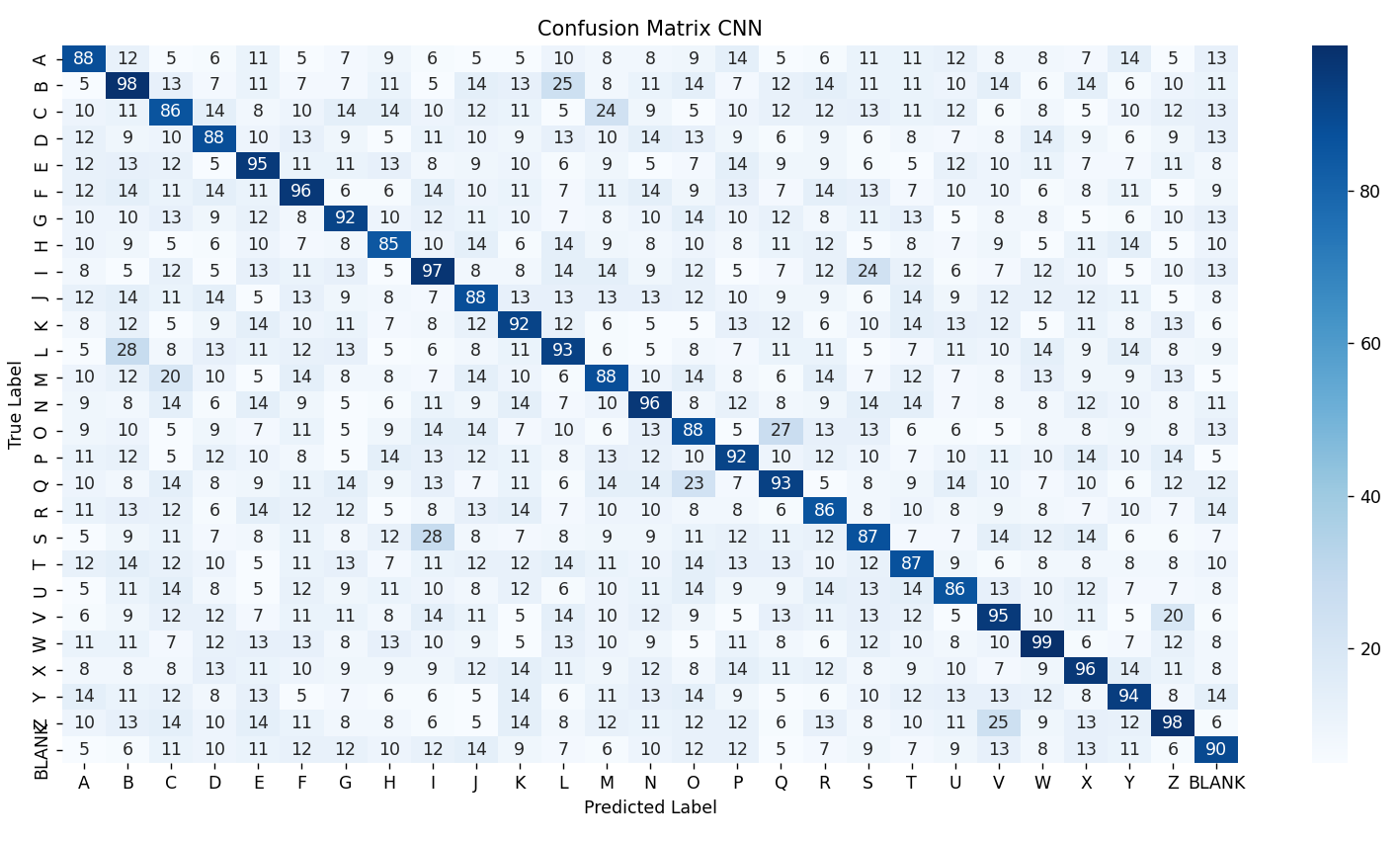}
    \caption{Confusion matrix Convolution Neural Network Model}
    \label{fig:conf_matrix_cnn}
\end{figure}

\subsection{Large Language Model}
For the text correction phase, the Fine-tuned Gemini-1.5 flash model was also validated against a sample set of 100 records it shows proper auto-correction with an accuracy of 94.2\% on the validation set.
% \begin{figure}[h]
%     \centering
%     \includegraphics[width=1.0\linewidth]{Images/fine_tuning_loss_epoch_llm.png}
%     \caption{Loss per epoch - Fine Tuning Gemini-1.5 flash}
%     \label{fig:fine_tune_le_g}
% \end{figure}

\begin{table}[h]
\centering
    \caption{Sample of tested inputs from validation set}
\label{tab:llm_comparison}
\begin{tabular}{ll}
\hline
\multicolumn{1}{c}{\textbf{Input}} & \multicolumn{1}{c}{\textbf{Output}}                                                             \\ \hline \hline
CENCICLOEDIA IE A BOOL & \begin{tabular}[c]{@{}l@{}}{[}'ENCYCLOPEDIA IS A BOOK',\\  'ENCYCLOPEDIA OF A BANK',\\  'A BOOK IS ENCYCLOPEDIA'{]}\end{tabular} \\
ABBLE STY ELTY         & \begin{tabular}[c]{@{}l@{}}{[}'EAT APPLE STAY HEALTHY',\\  'APPLE STAY THERE',\\  'APPLE IS HEALTHY'{]}\end{tabular}             \\
TOY BOK                            & \begin{tabular}[c]{@{}l@{}}{[}'TOY BOOK',\\  'BOOK OF TOYS',\\  'TOYS ARE BOOK'{]}\end{tabular} \\
MOVIE GOO                          & \begin{tabular}[c]{@{}l@{}}{[}'MOVIE GOOD',\\  'GOOD MOVIE',\\  'MOVIE IS GOOD'{]}\end{tabular} \\ \hline
\end{tabular}
\end{table}

\subsection{Video Outputs}

In the video synthesis phase, we get the following output after conversion from American sign language to Indian sign language on mapping.
\begin{figure}[h]
    \centering
    \includegraphics[width=0.65\linewidth]{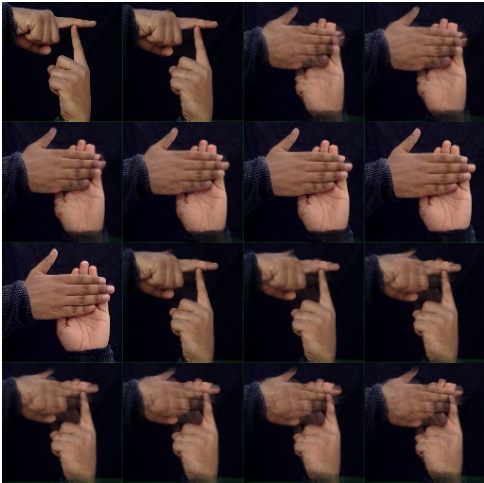}
    \caption{Frame interpolation using RIFE-Net}
    \label{fig:frame_inter}
\end{figure}
\\
The following figure shows the frames of the video output in indian sign language representing "THE BALL IS ON THE TABLE".
\begin{figure}[h]
    \centering
    \includegraphics[width=0.65\linewidth]{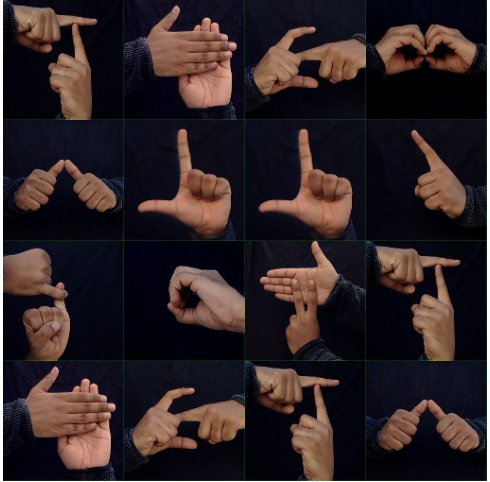}
    \caption{Video Output (1FPS)}
    \label{fig:vid_out}
\end{figure}
\section{Conclusion}
This research presents a novel framework for ASL-to-ISL translation, integrating advanced deep learning models, transfer learning, and large language models to bridge gaps in sign language communication. The system addresses linguistic and grammatical differences between ASL and ISL while ensuring context retention and culturally sensitive translation, facilitating interaction among diverse sign language users.

A key innovation is the ensemble model combining custom Convolutional Neural Networks (CNNs) and a Random Forest Classifier for robust ASL gesture recognition. This hybrid approach enhances recognition accuracy and captures nuanced gesture variability. High-end language models like Gemini 1.5 Flash, GPT-3.5, and LLaMA 2 (7B) are employed to translate gestures into text, preserving context and intent. Additionally, RIFE-Net enables real-time reconstruction of ISL gestures, resolving issues of gesture variability and temporal recovery.

Beyond addressing linguistic and technological challenges, the framework promotes inclusivity, enabling users with hearing or speech impairments to express themselves more effectively. This system offers significant potential for advancing accessible technologies and can serve as a foundation for future improvements, including incorporating other sign language dialects and multimodal features such as facial expressions and body language. Moreover, the system's adaptability, through user feedback and self-learning capabilities, can enhance precision and flexibility over time, contributing to the unification of global sign language use.

\section{Future Scope}
% or

This research lays a really good foundation for real-time translation from American Sign Language into Indian Sign Language is using state-of-the-art deep learning techniques and Large Language Models. As it handles static gestures and focuses on differences between ASL and ISL, the scope for this framework is huge, here are some areas that should be looked into and developed further:

\subsection{Expanding Support for Complete ISL and Other Sign Languages}

Even though this present system draws mostly from ISL, its sister signs differ significantly across regions and cultures, which has always led to one particular local sign including different grammatical structures and expressions. It may become worthy to include additional local versions in this work, like BSL (British Sign Language), LSF (French Sign Language), or ArSL (Arabic Sign Language). \\
\textbf{Challenges:} The vocabulary, regional dialects and non-standard formats vary with different sign languages. \\
\textbf{Implementation:} This can be achieved by training on diverse datasets that come with multi-regional sign languages. Fine-tuning LLMs based on a multilingual corpus enables them to effectively handle the nuances of cross-linguistic, thus making the framework adaptable to a global scale.

\subsection{Incorporating Dynamic Gestures}
Current theory is too focused on the static nature of translation, and most sign languages require dynamic gestures, which involve motion and transition over time.
\\
\textbf{Challenges:}  Recognition and interpretation of dynamic gestures require advanced qualitative as well as quantitative time analysis and the ability to distinguish nuances of movement. \textbf{Implementation:} Dynamic gestures can be dealt with processing systems like video analysis and processing that can be built using deep models like Transformer. Even further sequences of learning models may be utilized, such as Long Short-Term Memory or 3D convolutional networks, to get better insights into gesture transition.

\subsection{Enhancing Emotional Context Understanding}
Emotions are important in effective communication. Sign languages include facial expressions body language and tone of gestures in passing emotions, which the system doesn't fulfil to its potential.
\\
\textbf{Challenges:} Emotional context will be identified only by multimodal analysis, starting with facial expression and gesture intensity and posture. \\
\textbf{Implementation:} Future directions include multimodal input streams where gesture data are combined with facial emotion recognition systems based on convolutional or hybrid neural networks. Further training of the model on datasets annotated with emotional cues could enhance the system's capability of more holistically capturing the speaker's intent.

\subsection{Improved Real-Time Performance}
Real-time translation is decidedly computationally intensive, especially with the LLMs and high-fidelity gesture synthesis. This remains a critical area of improvement: speed and efficiency do not have to interfere with accuracy.
\\
\textbf{Challenges:} Balancing computational load with translation accuracy, especially on edge devices or resource-constrained environments. \\
\textbf{Implementation:} Optimization techniques applied include pruning, quantization, and adaptations to edge computing. Distributed processing or GPU acceleration of inference adds additional layers of real-time performance.
\subsection{ Contextual and Idiomatic Translation}
Idioms and context-specific meanings, which are valuable to sign languages, cannot be translated easily. There is some limitation in the usage of idioms in this system, as it is meant to preserve contextual integrity.
\\
\textbf{Challenges:} Capturing idiomatic context requires a deep understanding of cultural nuances and linguistic patterns.
\\
\textbf{Implementation:} Fine-tune LLMs with datasets that are rich in idiomatic phrases and also with reinforcement learning where feedback from the user can help fill in this gap. Adaptive learning algorithms user-centric will make translations over time more sensitive to the context.

\subsection{Interactive Feedback Mechanism}
The future system may even comprise an interactive feedback loop where users can modify or even validate the translations. This loop of continuous learning would, in that case, therefore improve precision and responsiveness with time.
\\
\textbf{Implementation:} A reinforcement learning module implementation based on user feedback using the improvement of the prediction model helps in effective gesture recognition and translation processes.

% \appendices
% Appendix One text goes here.

% % You can choose not to have a title for an appendix
% % if you want by leaving the argument blank
% \section{}
% Appendix Two text goes here.

% \section*{Acknowledgment}
% The authors would like to express their sincere gratitude to Dr. Anisha Natarajan, from the School of Electronics Engineering, Vellore Institute of Technology, Chennai, for her invaluable guidance, support, and mentorship throughout this research. We also extend our heartfelt thanks to the School of Computer Science and Engineering, Vellore Institute of Technology, Chennai, for providing the necessary resources and an enriching academic environment that facilitated the successful completion of this work.

\ifCLASSOPTIONcaptionsoff
  \newpage
\fi

\end{document}